\documentclass{article}
\usepackage{spconf,amsmath,graphicx}
\usepackage{amssymb}
\usepackage{multirow}
\usepackage{multicol}
\usepackage{array}
\newcolumntype{P}[1]{>{\centering\arraybackslash}p{#1}}
\usepackage{hyperref}
\hypersetup{
    colorlinks=true,
    linkcolor=blue,
    filecolor=magenta,      
    urlcolor=black,
}
\title{Deep Laplacian Pyramid Network for Text Images Super-Resolution}
\name{Hanh T. M. Tran, Tien Ho-Phuoc}
\address{The University of Danang - University of Science and Technology
\\ Email: t2mhanh@gmail.com, hptien@dut.udn.vn}

\begin{document}
%
\maketitle
\begin{abstract}
Convolutional neural networks have recently demonstrated interesting results for 
single image super-resolution. However, these 
networks were trained to deal with super-resolution problem on natural images. In this paper, we adapt a deep network, which was proposed for natural images super-resolution, to single text image super-resolution. To evaluate the network, we present our database for single text image super-resolution. Moreover, we propose to combine Gradient Difference Loss (GDL) with L1/L2 loss to enhance edges in super-resolution image. Quantitative and qualitative evaluations on our dataset show that adding the GDL improves the super-resolution results.  
\end{abstract}
\begin{keywords}
Deep Laplacian Pyramid Networks, Convolutional neural networks, Text Images, Super-Resolution. 
\end{keywords}
\section{Introduction}
\label{sec:intro}
Super resolution (SR) is an important topic in image processing and computer vision thanks to its ability to generate pleasing images and facilitate the tasks of detection and recognition~\cite{bulat08}. 

According to the input of the SR process, we can divide SR into two categories: multiple image super resolution (MISR) and single image super resolution (SISR). Generally, MISR exploits different information in neighboring frames to reconstruct a high resolution (HR) image. This approach may require explicit or implicit motion estimation~\cite{farsiu04,hophuoc13}.

Recently, more researches seem focused on SISR as it allows fast and efficient reconstruction of HR images. However, this is a highly ill-posed problem. Except simple methods, e.g. bicubic interpolation, SISR needs a learning step, which may learn a sparse representation in the low resolution (LR) and high resolution spaces \cite{yang08}, or a mapping function from the LR space to the HR space~\cite{timofte08}. Besides, inspired by the success of convolutional neural network (CNN) in high-level tasks such as object recognition, many authors have tried to apply CNN to the SR problem and have obtained very interesting results~\cite{dong14,lai2017deep}. 
Until now, a large majority of SR researches, and therefore their success, have been carried out with natural images. Yet, few works have dealt with text image SR, and particularly, single text image SR \cite{dalley04, banerjee08, nayef14, dong15,xu17}; although this topic can be very useful for many applications such as optical character recognition (OCR), text detection, and scene understanding.

Motivated by the potential of CNN, the present paper focuses on super resolution of single text images using a deep network, which is adapted from Lai's model~\cite{lai2017deep}, originally proposed for natural images. Although our method is based on Lai's model \cite{lai2017deep}, we also make the following contributions. Firstly, we modify the L1/L2 loss function by adding Gradient Difference Loss (GDL) \cite{mathieu16} in order to sharpen edges, which are critical in text images. Secondly, different from previous works on SR of text images, we consider multiple SR scales. Finally, we present a new dataset for text image super resolution. This dataset may be useful for SR in a general context since it contains not only paragraphs but also words or sentences in banners and newspaper.  

\section{Related Work}
\textbf{Text image super resolution.} 
Multiple-frame SR for text images are addressed in~\cite{capel00,nasonov12}. Capel et al.~\cite{capel00} utilize a MAP approach with Huber prior, which encourages a piecewise constant solution. The authors compare this regularization with Total Variation, the latter is proved to be of easier use in practice. Nasonov et al.~\cite{nasonov12} also use Total Variation, but adding a bimodal penalty function, which represents text background and text foreground. Both these works deal only with a 2x upscaling ratio.

Concerning single-frame text super resolution, Dalley et al.~\cite{dalley04} formulate a Bayesian framework to super-resolve binary text images, aiming at improving the visual quality of fax documents or low resolution scans. Banerjee et al.~\cite{banerjee08} try to preserve edges while super-resolving grayscale document images. Edge preservation is enforced through a Markov Random Field framework.

Nayef et al.~\cite{nayef14} propose to super-resolve an image patch according to its variance. If a patch has low variance, a fast bicubic interpolation is used. Otherwise, sparse coding SR is applied. Originally, SR based on sparse coding or sparse representation is proposed in~\cite{yang08}. According to this approach, a LR image is first represented in the LR space by a sparse set of coefficients and a LR dictionary; these coefficients are then used with a HR dictionary to reconstruct a HR image. The method in \cite{nayef14} super-resolves images of plain text documents extracted from books.

\textbf{Convolutional neural network.} 
The first CNN for super resolution (SRCNN) is proposed in~\cite{dong14}. In this method, a LR image is first upsampled to the predefined HR scale by bicubic interpolation. Then, the resulting image is super-resolved (in fact, sharpened) by a network with three conv-ReLU layers (except that the last convolutional layer is not followed by a ReLU layer). It is interesting to note that although SRCNN is formulated in the CNN context, it is proved to be equivalent to learning a sparse representation in the LR and HR spaces. Since SRCNN's success, many other authors have tried to propose various CNN architectures, particularly increase network depth, and significantly improve SR performance~\cite{wang15,lai2017deep}.

Among the recent CNNs for super resolution, the Laplacian pyramid network (LapSRN) \cite{lai2017deep} seems to be the state-of-the-art for natural images. The particularity of this method is to directly generate a HR image from a single LR input, this is done thanks to a pyramidal processing pipeline with several stages. Each stage, consisting of ten convolutional layers mainly to effectuate residual learning, carries out a 2x upscaling ratio. While our network is similar to LapSRN \cite{lai2017deep}, we utilizes a different loss function by combining L1/L2 loss with Gradient Difference Loss (GDL). The intuitive idea of using GDL is that this constraint might sharpen letter edges and, therefore, result in clearer text. 

Concerning CNN for text data, in \cite{dong15}, the authors adapt SRCNN \cite{dong14} for text images in order to improve the OCR performance. The dataset includes small grayscale images, each of which contains only few words or numbers. Xu et al.~\cite{xu17} use a generative adversarial network (GAN) to super-resolve both blurry face and text images. The dataset of text images is extracted from scientific publications. This dataset is also exploited in \cite{hradis15}, where a CNN is used for text deblurring, but not for super resolution.  

Our work tries to super-resolve color text images. Different from previous text datasets, these images represent text in a broader context as they include not only paragraphs but also words and sentences in banners or complex background. Super-resolving such kinds of text images may be useful for text detection and recognition in real scenes. Moreover, we consider multiple upscaling ratios; such scenario has not been examined for text image super resolution. 

\section{Deep Laplacian Pyramid Networks for super-resolution}
\subsection{Network Architecture}
\begin{figure}[t]
\begin{center}
\includegraphics[width=0.52\textwidth]{{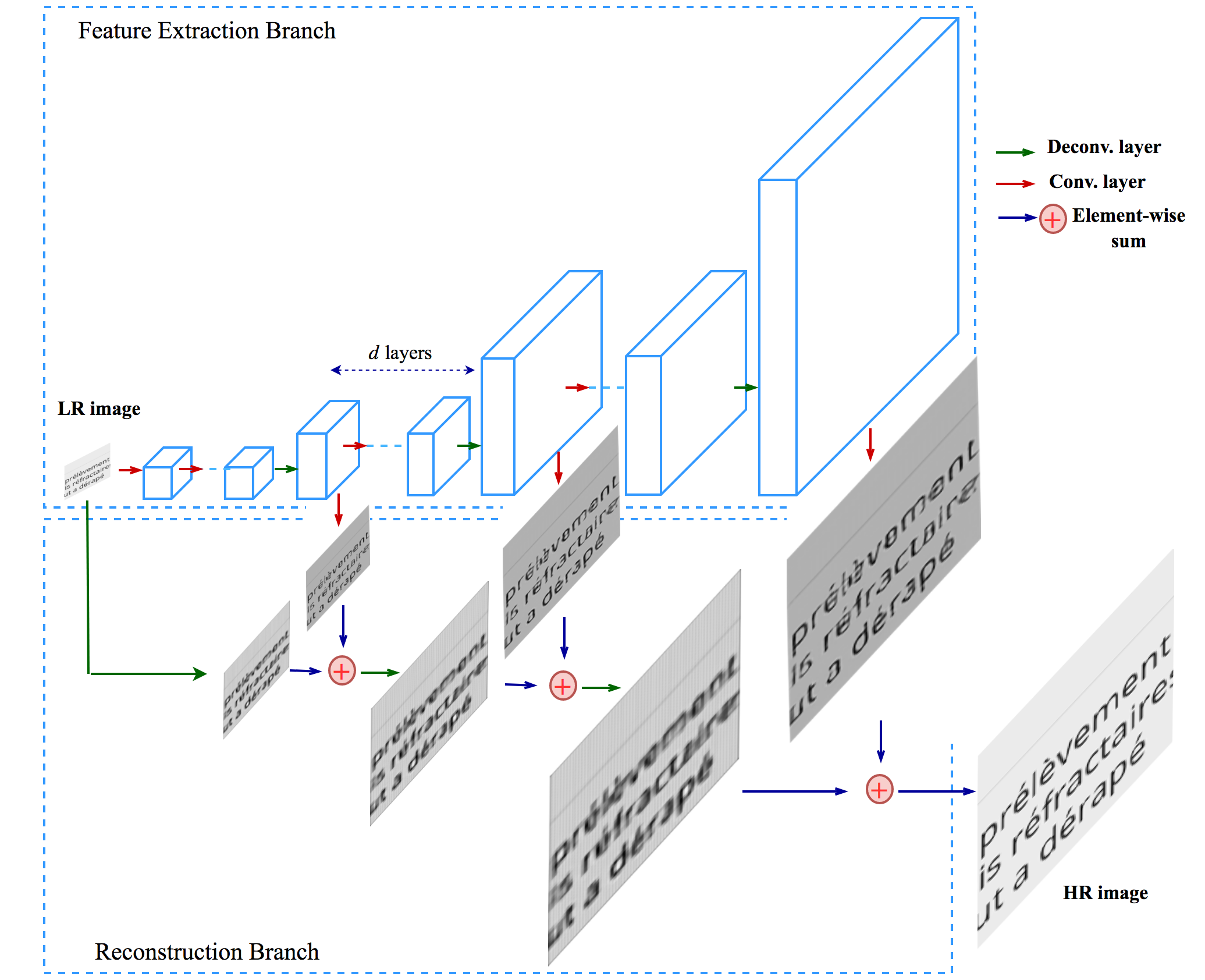}} 
\end{center}
   \caption{Network architecture of Laplacian Super-resolution on text images.}
\label{fig:architecture_x8}
\end{figure}

We employ the same network as in the Laplacian pyramid framework \cite{lai2017deep}, as shown in Figure \ref{fig:architecture_x8}. The model takes an LR image as input and progressively predicts images at log$_2S$ levels where $S$ is a scale factor. The model has two branches: feature extraction and image reconstruction. 

At level $s$, the feature extraction branch consists of $d$ convolutional layers following by leaky ReLU layers and one deconvolutional layer to upsample the extracted features by a scale of 2. The output of the deconvolutional layer is fed into two different layers: (1) a convolutional layer for extracting features at the finer level $s+1$, and (2) a convolutional layer for reconstructing a residual image at level $s$ which is then combined with upsampled LR image in the reconstruction branch.    

The image reconstruction branch consists of a deconvolutional layer at level $s$ to up-sample an input image by a scale of $2$. 
 The upsampled image is combined with the predicted residual image from the feature extraction branch to produce a high-resolution output image. The output image at level $s$ is then fed into the image reconstruction branch of level $s+1$. 
 
 We train three networks to produce $2\times, 4\times$ and $8\times$ results. At each level, number of convolutional layers $d=10$ is used for all networks. Figure \ref{fig:architecture_x8} shows the network for $8 \times $ super-resolution. $8 \times$ model can be used to produce $2 \times$ and $4 \times$ images. 
\begin{figure*}[!t]
\begin{center}
\begin{tabular}{c c c}
\includegraphics[width=0.3\textwidth]{{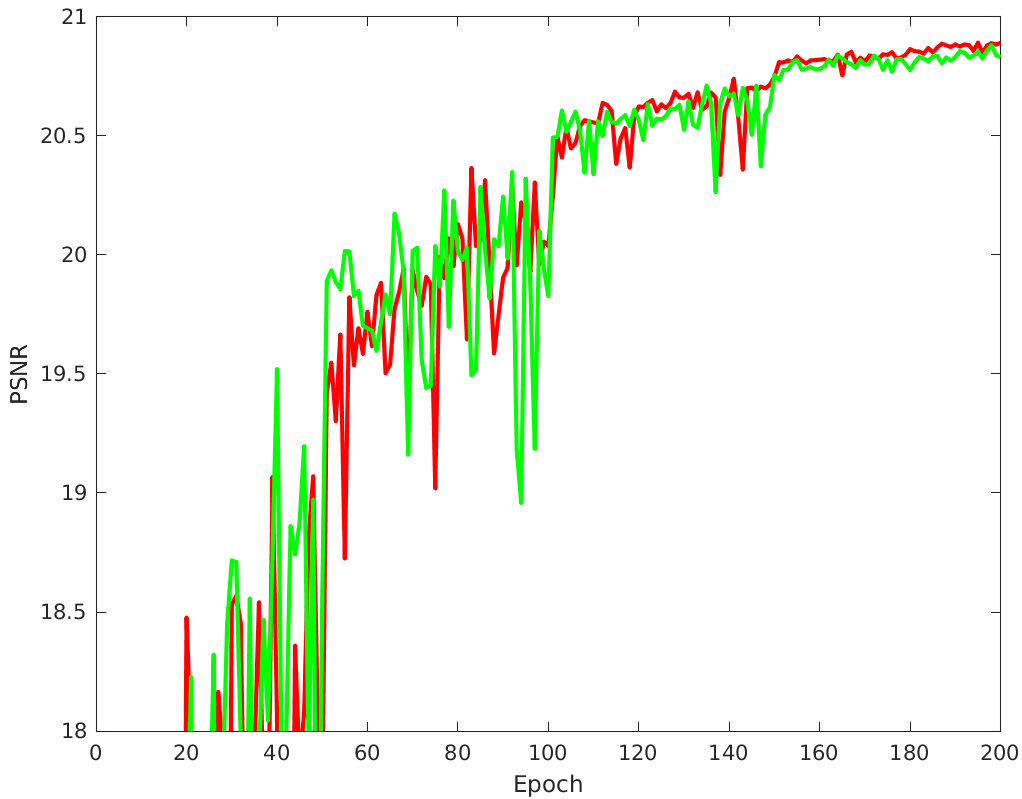}} & \includegraphics[width=0.3\textwidth]{{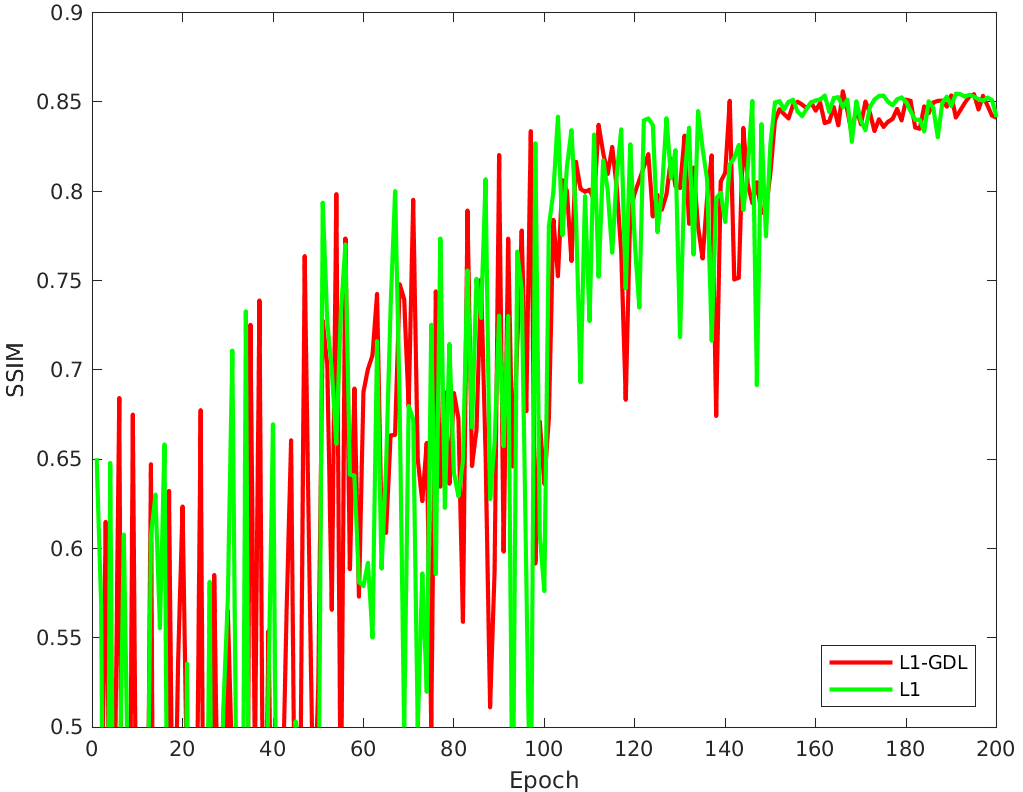}} & \includegraphics[width=0.3\textwidth]{{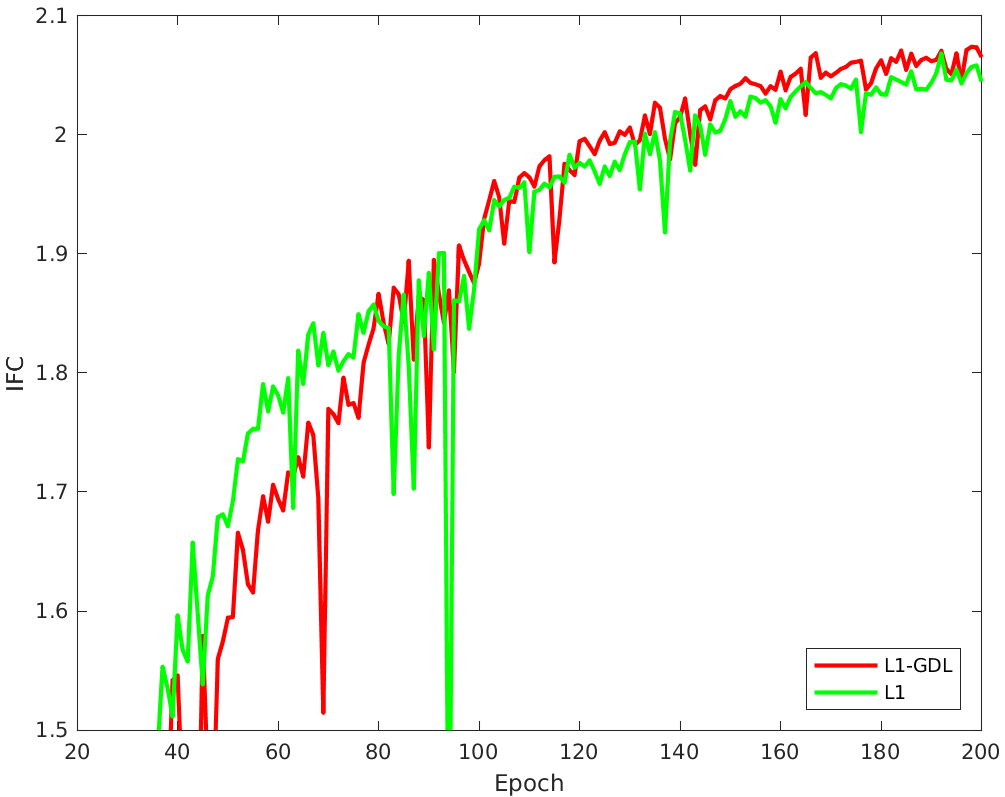}}\\
\end{tabular}
\end{center}
   \caption{PSNR, SSIM and IFC measurements over number of training epochs.}
\label{fig:PSNR_epochs}
\end{figure*}
\subsection{Loss function}
As in \cite{lai2017deep}, our goal is to learn a mapping function $f$ for generating an HR image $\hat{\mathbf{Y}} = f(\mathbf{X},\pmb{\theta})$ from a LR image $\mathbf{X}$ that is close to the ground truth HR image $\mathbf{Y}$. We denote the HR image at level $s$ by $\mathbf{Y}_s$. The bicubic dowsampling is used to resize the ground truth image $\mathbf{Y}$ to $\mathbf{Y}_s$ at each level. 
Beside the use of Charbonnier penalty function \cite{lai2017deep}, we also combine the Gradient Difference Loss \cite{mathieu16} to sharpen the estimation $\hat{\mathbf{Y}}$of the HR image. 
\begin{equation} \label{eq:loss2}
\begin{split}
    \mathcal{L}_{gdl}(\mathbf{Y},\hat{\mathbf{Y}}) & = \sum_{i,j}\rho(|\mathbf{Y}_{i,j} - \mathbf{Y}_{i-1,j}| - |\hat{\mathbf{Y}}_{i,j} - \hat{\mathbf{Y}}_{i-1,j}|) +\\
    & \sum_{i,j}\rho(|\mathbf{Y}_{i,j-1} -  \mathbf{Y}_{i,j}| - |\hat{\mathbf{Y}}_{i,j-1} - \hat{\mathbf{Y}}_{i,j}|)    
\end{split}
\end{equation}
The network parameter $\pmb{\theta}$ are learned by minimizing the following loss function: 
\begin{equation} \label{eq:loss1}
\mathcal{L}(\mathbf{Y},\hat{\mathbf{Y}}) = \frac{1}{N}\sum_{n=1}^N \sum_{s=1}^L (\rho(\hat{\mathbf{Y}}_s^{(n)} - \mathbf{Y}_s^{(n)}) + \lambda_{gdl} \mathcal{L}_{gdl}(\mathbf{Y}_s^{(n)},\hat{\mathbf{Y}}_s^{(n)}))
\end{equation}
where $\rho(x) = \sqrt{x^2 + \epsilon^2}$ is the Charbonnier penalty function (a variant of L$_1$ norm), $N$ is the number of training samples in each batch and $L$ is the number of levels in the pyramid. We empirically set $\epsilon = 1e-3$. $\lambda_{gdl}$ is the weight to emphasize the GDL loss in the total loss.

\subsection{Training}
In the network, each convolutional layer consists of 64 filters with the size of $3 \times 3$. 
The weights in each convolutional layer are initialized from a zero-mean Gaussian distribution whose standard deviation is calculated from the number of input channels and the spatial filter size of the layer~\cite{he2015delving}. This is a robust initialization method that particularly considers the rectifier nonlinearities.  The size of the deconvolutional layers filters is $4 \times 4$ and the weights are initialized with the bilinear kernel. 

We use SGD with momentum to optimize the error in Eq. \ref{eq:loss1} with batch size \(N=64\), momentum of $0.9$ and $0.999$, weight decay  \(\lambda = 10^{-4}\) ~\cite{krizhevsky2012imagenet}. We start training the model with learning rate of $10^{-5}$ and decrease a half after 50 epochs. We train models on a GeForce GTX Titan X around two days. Code and trained models are available at \textit{\url{https://github.com/t2mhanh/LapSRN_TextImages_git}}.  

Similar to \cite{lai2017deep}, we augment the training data in three methods: scaling, rotation and flipping. Moreover, we use patches size of $128\times128$ as HR outputs of the models. LR training patches are generated by bicubic downsampling on HR patches. The models are trained with Matconvnet toolbox \cite{vedaldi2015matconvnet}. 
\section{Experimental Results}
\subsection{Database}
While existing text image datasets for super resolution often concern only binary or grayscale images and plain text documents; we propose a new dataset (Text330) that considers text images in a broader and more realistic context. They are extracted from ebooks, websites, and newspaper. The text is in three languages - English, French, and Vietnamese - in order to take into account language particularities. Concretely, beside plain text images such as paragraphs, the proposed dataset also consists of numbers, characters, and sentences in complex color background. The text may appear along with persons or natural scenes as we can see in newspaper and banners. Font size may vary within and between images; font style is normal or italic. 

In total, there are 330 images divided into two parts: 300 images for training and 30 images for testing. In each part, the number of images is equally partitioned in the three languages.       
\begin{figure*}[!ht]
\begin{center}

\begin{tabular}{|P{5.5cm}| P{2.2cm}| P{2.2cm} |P{2.2cm} |P{2.2cm}| }
\hline
Ground-truth HR & Bicubic & Pre-trained model & L1 loss & L1-GDL \\
\hline
\includegraphics[width=0.3\textwidth]{{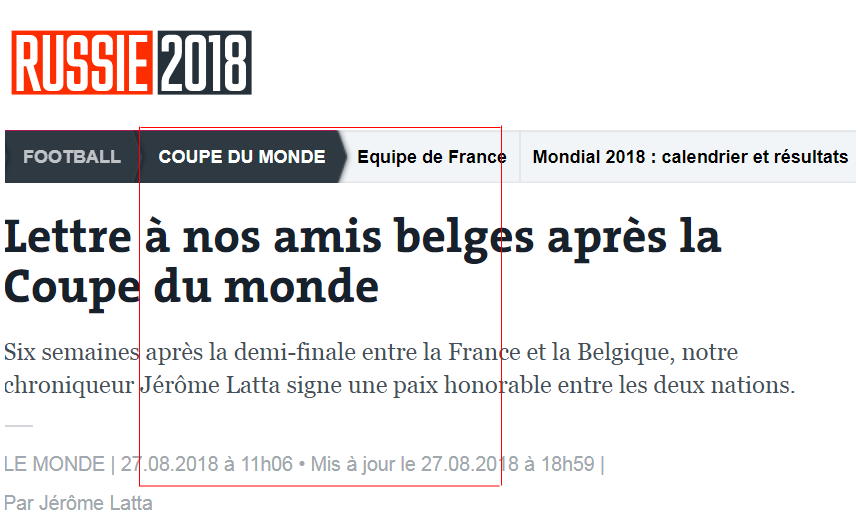}} & \includegraphics[width=0.12\textwidth]{{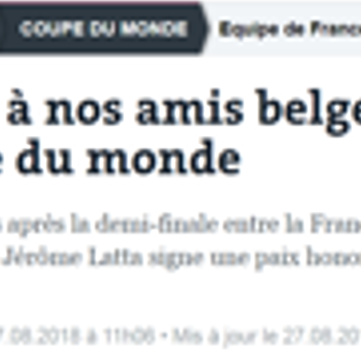}} & \includegraphics[width=0.12\textwidth]{{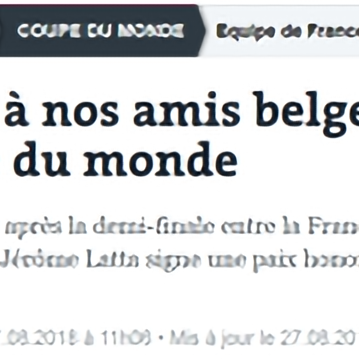}} & \includegraphics[width=0.12\textwidth]{{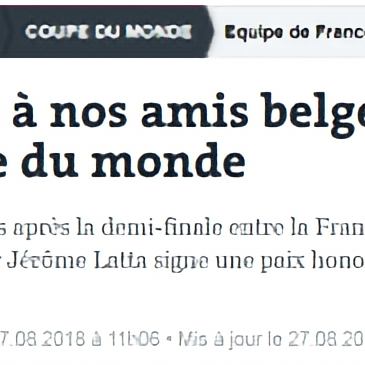}} & \includegraphics[width=0.12\textwidth]{{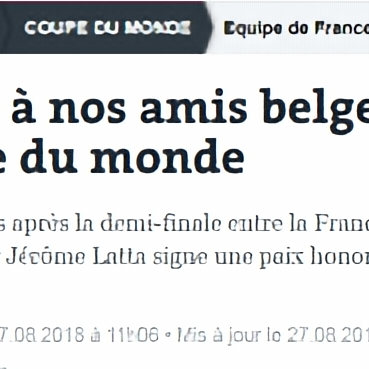}}\\
\scriptsize (PSNR, SSIM) & \scriptsize ($21.01,0.8393$)  & \scriptsize ($23.55,0.9256$) &  \scriptsize ($26.03,0.9623$) & \scriptsize ($\mathbf{26.13},\mathbf{0.9635}$) \\
 \hline
\includegraphics[width=0.3\textwidth]{{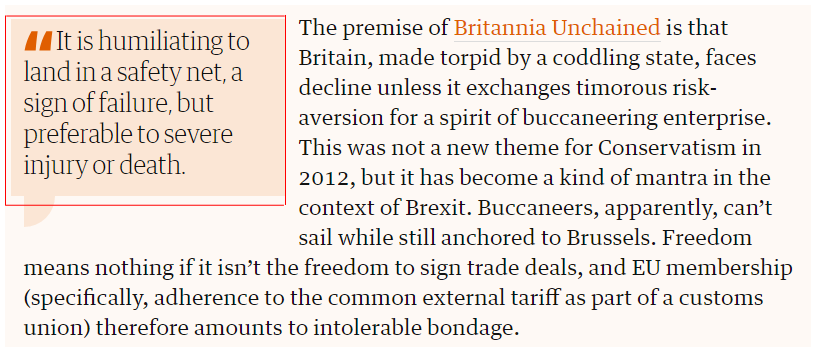}} & \includegraphics[width=0.12\textwidth]{{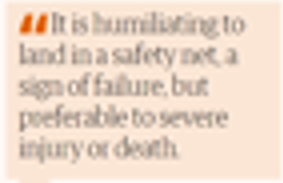}} & \includegraphics[width=0.12\textwidth]{{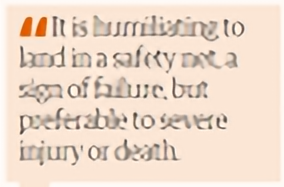}} & \includegraphics[width=0.12\textwidth]{{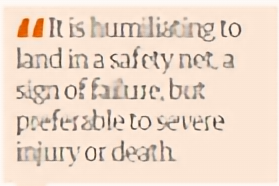}} & \includegraphics[width=0.12\textwidth]{{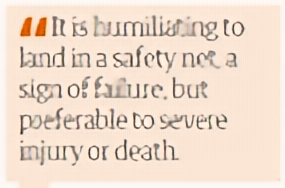}}\\
         & \scriptsize ($16.35,0.5895$)  & \scriptsize ($17.09,0.7339$) &  \scriptsize ($\mathbf{20.24},\mathbf{0.8704}$) &  \scriptsize ($20.19,0.8698$) \\
  \hline
\includegraphics[width=0.3\textwidth]{{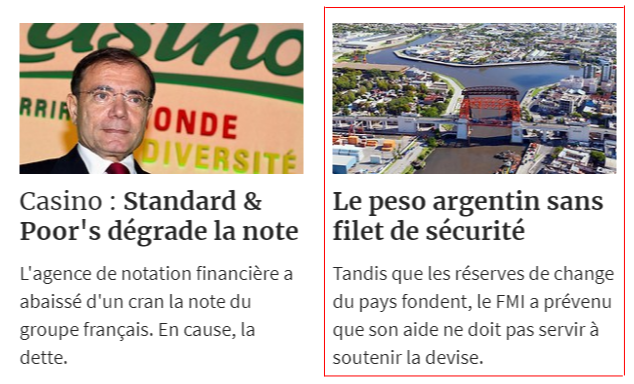}} & \includegraphics[width=0.12\textwidth]{{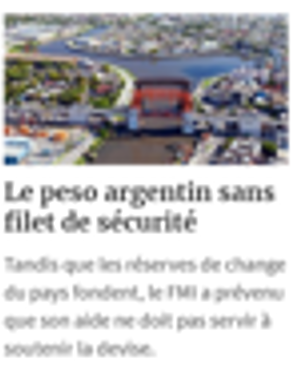}} & \includegraphics[width=0.12\textwidth]{{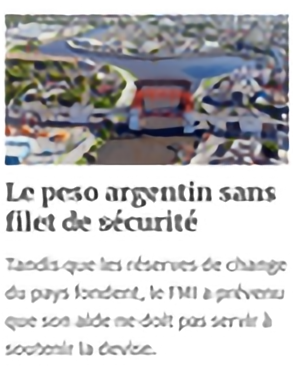}} & \includegraphics[width=0.12\textwidth]{{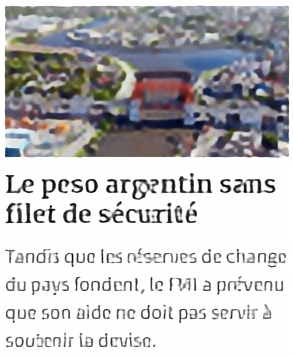}} & \includegraphics[width=0.12\textwidth]{{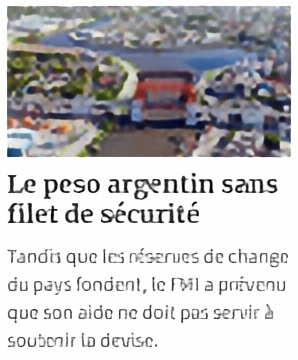}}\\
          & \scriptsize ($18.68,0.6581$)  & \scriptsize ($19.92,0.7762$) & \scriptsize ($22.01,0.8416$) & \scriptsize ($\mathbf{22.17},\mathbf{0.8437}$) \\
 \hline
\includegraphics[width=0.3\textwidth]{{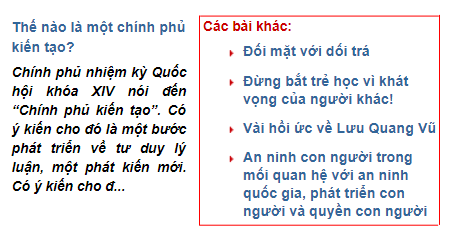}} & \includegraphics[width=0.12\textwidth]{{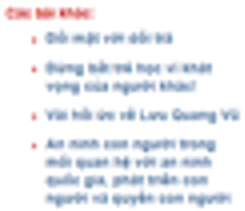}} & \includegraphics[width=0.12\textwidth]{{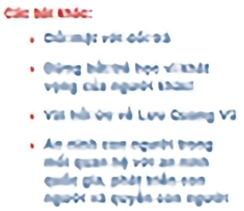}} & \includegraphics[width=0.12\textwidth]{{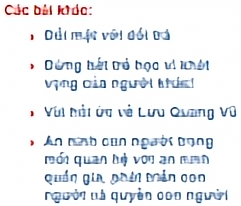}} & \includegraphics[width=0.12\textwidth]{{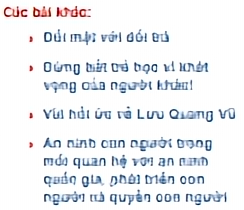}}\\
         & \scriptsize ($17,0.6546$)  & \scriptsize ($17.77,0.7594$) &  \scriptsize ($18.95,0.8556$) &  \scriptsize ($\mathbf{19.9},\mathbf{0.86}$) \\
\hline
\end{tabular}
\end{center}
   \caption{Visual comparison with (PSNR, SSIM) for $4 \times$ SR on our dataset.}
\label{fig:visual_comparison}
\end{figure*}
\subsection{Evaluation method}
Three commonly used performance metrics are employed for evaluation and comparison: PSNR, SSIM \cite{wang2004image}, IFC \cite{sheikh2005information}. They are all based only on luminance. The higher these criteria, the better the quality of the reconstructed image.

We will describe the three criteria as follows. Let $\mathbf{Y}$ and $\hat{\mathbf{Y}}$ denote the reference (ground truth) and reconstructed images respectively.
\begin{equation} \label{eq:PSNR}
\textrm{PSNR}(\hat{\mathbf{Y}},\mathbf{Y}) = 10 \log_{10} \frac{255^2}{\textrm{MSE}}
\end{equation}
with 
\begin{equation} \label{eq:MSE}
\textrm{MSE} =\frac{1}{MN} \sum_{i=1}^M \sum_{j=1}^N (\hat{\mathbf{Y}}(i,j)-\mathbf{Y}(i,j))^2
\end{equation}

The SSIM criterion between patches $\mathbf{P_{\hat{\mathbf{Y}}}}$ and $\mathbf{P_Y}$ at the same location on images $\hat{\mathbf{Y}}$ and $\mathbf{Y}$ is defined as
\begin{equation} \label{eq:SSIM}
\textrm{SSIM}(\mathbf{P_{\hat{\mathbf{Y}}}},\mathbf{P_Y}) = \frac{(2\mu_{\mathbf{P_{\hat{\mathbf{Y}}}}}\mu_{\mathbf{P_Y}}+c_1)(2\sigma_{\mathbf{P_{\hat{\mathbf{Y}}}}} \sigma_{\mathbf{P_Y}}+c_2)}{(\mu_{\mathbf{P_{\hat{\mathbf{Y}}}}}^2+\mu_{\mathbf{P_Y}}^2+c_1)(\sigma_{\mathbf{P_{\hat{\mathbf{Y}}}}}^2+\sigma_{\mathbf{P_Y}}^2+c_2)}
\end{equation}
where $\mu_{\mathbf{P_{\hat{\mathbf{Y}}}}}$ ($\mu_{\mathbf{P_Y}}$) and $\sigma_{\mathbf{P_{\hat{\mathbf{Y}}}}}$ ($\sigma_{\mathbf{P_Y}}$) are the mean and standard deviation of patch $\mathbf{P_{\hat{\mathbf{Y}}}}$ ($\mathbf{P_Y}$). 
$c_1$ and $c_2$ are small constants. Then, SSIM($\hat{\mathbf{Y}}$,$\mathbf{Y}$) is the average of patch-based SSIM over the image.

The IFC criterion is based on the conditional mutual information:
\begin{equation} \label{eq:IFC}
\textrm{IFC}(\hat{\mathbf{Y}},\mathbf{Y}) =\sum_{k}I(C_{\hat{\mathbf{Y}},k};C_{\mathbf{Y},k}|s_k) 
\end{equation}
where $C_{\hat{\mathbf{Y}},k}$ ($C_{\mathbf{Y},k}$) is the set of wavelet coefficients of $\hat{\mathbf{Y}}$ ($\mathbf{Y}$) in subband $k$; and $s_k$ is a random field of positive scalars and is estimated from reference image $\hat{\mathbf{Y}}$. The details of IFC implementation can be found in \cite{sheikh2005information}. 

However, it is important to note that objective evaluation such as PSNR, SSIM, and IFC is not enough, and, usually, we still need humans for a complete evaluation.
\begin{table*}[h]
\caption{Performance comparison with different methods.}
\label{table:comp1}
\begin{center}
\begin{tabular}{l p{1.2cm} p{1.2cm} p{1.2cm} p{1.2cm}}
\hline
Algorithm & Scale & PSNR & SSIM & IFC\\
\hline
Bicubic & 2	& 21.82 & 0.869 & 3.378 \\
LapSRN pre-trained model \cite{lai2017deep} & 2	& 26.65	& 0.954 & 5.284\\
LapSRN L1-GDL   & 2 & $\mathbf{30.10}$ & $\mathbf{0.997}$ & $\mathbf{6.659}$ \\
LapSRN L1  & 2 & 29.98  & 0.976 & 6.590\\
\hline
Bicubic & 4	& 18.21 & 0.692 & 1.189 \\
LapSRN pre-trained model \cite{lai2017deep} & 4	& 19.64	& 0.796 & 1.876\\
LapSRN L1-GDL & 4 & $\mathbf{21.62}$ & $\mathbf{0.875}$ & $\mathbf{2.334}$ \\
LapSRN L1 & 4 & 21.55 & 0.873 & 2.296\\
\hline
Bicubic & 8	& 16.53 & 0.570 & 0.407 \\
LapSRN pre-trained model \cite{lai2017deep} & 8	& 16.98	& 0.640 & 0.594\\
LapSRN L1-GDL 
& 8 & $\mathbf{17.14}$ & 0.651 & $\mathbf{0.801}$\\
LapSRN L1 
& 8 & 17.13 & $\mathbf{0.682}$ & 0.788 \\
\hline
\end{tabular}
\end{center}
\end{table*}
%
\begin{table}[h]
\caption{L1-GDL with different values of $\lambda_{gdl}$ ($4 \times$ super-resolution).}
\label{table:lambda_gdl}
\centering
\begin{tabular} {l p{2.5cm} p{1.1cm} p{1.1cm} p{1.1cm}}
\hline
$\lambda_{gdl}$ & $lr$ range &  PSNR & SSIM & IFC \\
\hline
0 & $10^{-5} \div 10^{-6}$ & 21.55 & 0.873 & 2.296\\
0.05 & $5 \times 10^{-6} \div 10^{-6}$ & 21.44 & 0.866 & 2.291\\
$0.1$ & $10^{-5} \div 10^{-6}$ & $\mathbf{21.62}$ & $\mathbf{0.875}$ & $\mathbf{2.334}$ \\
0.5 & $10^{-6} \div 10^{-7}$ & 20.15 & 0.826 & 1.857 \\
$1$ & $10^{-6} \div 10^{-7}$ & 20.21 & 0.831 & 1.911 \\
\hline
\end{tabular}
\end{table}
%
\subsection{Comparisons with different methods}
We compare the method with different loss functions. Moreover, we also compare our results with bicubic interpolation method and LapSRN pre-trained models \cite{lai2017deep}. Table \ref{table:comp1} shows quantitative comparisons for $2\times$, $4\times$ and $8\times$ super-resolution. 
PSNR, SSIM and IFC are better with the combination of L1 (Chanonnier) loss and GDL. The results are improved significantly on $2 \times$ and $4 \times$ SR. Visual comparisons with (PSNR, SSIM) for $4 \times$ SR on our dataset are shown in Figure \ref{fig:visual_comparison}. As in the figure, the SR images are clearer on our models (without/with GDL) in comparison with bicubic interpolation and LapSRN pre-trained models. Table \ref{table:lambda_gdl} shows the results with different values of $\lambda_{gdl}$ in which $\lambda_{gdl} = 0.1$ give the best results. This value of $\lambda_{gdl}$ is used to train our models and to compare with bicubic interpolation and LapSRN pre-trained models.  
\section{Conclusion}
In this work, we propose to use a deep convolutional network for single text image super-resolution. The model progressively reconstructs high-frequency residuals in a coarse-to-fine manner. By adding Gradient Difference Loss, the high resolution estimation becomes sharper. Evaluations on our text image dataset demonstrate that the model performs better than bicubic interpolation and the pre-trained models. Moreover, the combination of L1 loss and GDL improves the super-resolution results.
\bibliographystyle{IEEEbib}
\bibliography{Hanh_IEEEexample}

\begin{thebibliography}{10}

\bibitem{bulat08}
Adrian Bulat and Georgios Tzimiropoulos,
\newblock ``Super-fan: Integrated facial landmark localization and
  super-resolution of real-world low resolution faces in arbitrary poses with
  gans,''
\newblock in {\em CVPR}, 2018, pp. 109--117.

\bibitem{farsiu04}
Sina Farsiu, Dirk Robinson, Michael Elad, and Peyman Milanfar,
\newblock ``Fast and robust multiframe super resolution,''
\newblock {\em IEEE transactions on image processing}, vol. 13, no. 10, pp.
  1327--1344, 2004.

\bibitem{hophuoc13}
T.~Ho-Phuoc, A.~Dupret, and L.~Alacoque,
\newblock ``Super resolution method adapted to spatial contrast,''
\newblock in {\em ICIP}, 2013, pp. 976--980.

\bibitem{yang08}
J.~Yang, J.~Wright, T.~Huang, and Y.~Ma,
\newblock ``Image superresolution as sparse representation of raw image
  patches,''
\newblock in {\em CVPR}, 2008.

\bibitem{timofte08}
R.~Timofte, V.~De Smet, and L.~Van Gool,
\newblock ``A+: Adjusted anchored neighborhood regression for fast
  super-resolution,''
\newblock in {\em ACCV}, 2014.

\bibitem{dong14}
C.~Dong, C.C. Loy, K.~He, and X.~Tang,
\newblock ``Learning a deep convolutional network for image super-resolution,''
\newblock in {\em ECCV}, 2014, pp. 184--199.

\bibitem{lai2017deep}
Wei-Sheng Lai, Jia-Bin Huang, Narendra Ahuja, and Ming-Hsuan Yang,
\newblock ``Deep laplacian pyramid networks for fast and accurate
  superresolution,''
\newblock in {\em IEEE Conference on Computer Vision and Pattern Recognition},
  2017, vol.~2, p.~5.

\bibitem{dalley04}
Gerald Dalley, Bill Freeman, and Joe Marks,
\newblock ``Single-frame text super resolution: a bayesian approach,''
\newblock in {\em ICIP}, 2004, pp. 3295--3298.

\bibitem{banerjee08}
J.~Banerjee and C.~Jawahar,
\newblock ``Super-resolution of text images using edge directed tangent
  field,''
\newblock in {\em International Workshop on Document Analysis Systems}, 2008,
  pp. 96--83.

\bibitem{nayef14}
Nibal Nayef, Joseph Chazalon, Petra Gomez-Krämer, and Jean-Marc Ogier,
\newblock ``Efficient example-based super-resolution of single text images
  based on selective patch processing,''
\newblock in {\em International Workshop on Document Analysis Systems}, 2014,
  pp. 227--231.

\bibitem{dong15}
C.~Dong, X.~Zhu, Y.~Deng, C.~C. Loy, and Y.~Qiao,
\newblock ``Boosting optical character recognition: A super-resolution
  approach,''
\newblock in {\em International Conference on Document Analysis and
  Recognition}, 2015.

\bibitem{xu17}
X.~Xu, D.~Sun, J.~Pan, Y.~Zhang, H.~Pfister, and M.~H. Yang,
\newblock ``Learning to super-resolve blurry face and text images,''
\newblock in {\em ICCV}, 2017.

\bibitem{mathieu16}
Michael Mathieu, Camille Couprie, and Yann LeCun,
\newblock ``Deep multi-scale video prediction beyond mean square error,''
\newblock in {\em ICLR}, 2016.

\bibitem{capel00}
David Capel and Andrew Zisserman,
\newblock ``Super-resolution enhancement of text image sequences,''
\newblock in {\em ICPR}, 2000.

\bibitem{nasonov12}
Andrey~V. Nasonov and Andrey~S. Krylov,
\newblock ``Text images super resolution and enhancement,''
\newblock in {\em International Congress on Image and Signal Processing}, 2012.

\bibitem{wang15}
Z.~Wang, D.~Liu, J.~Yang, W.~Han, and T.~Huang,
\newblock ``Deep networks for image super-resolution with sparse prior,''
\newblock in {\em ICCV}, 2015.

\bibitem{hradis15}
Michal Hradis, Jan Kotera, Pavel Zemcik, and Filip Sroubek,
\newblock ``Convolutional neural networks for direct text deblurring,''
\newblock in {\em British Machine Vision Conference}, 2015.

\bibitem{he2015delving}
Kaiming He, Xiangyu Zhang, Shaoqing Ren, and Jian Sun,
\newblock ``Delving deep into rectifiers: Surpassing human-level performance on
  imagenet classification,''
\newblock in {\em Proceedings of the IEEE international conference on computer
  vision}, 2015, pp. 1026--1034.

\bibitem{krizhevsky2012imagenet}
Alex Krizhevsky, Ilya Sutskever, and Geoffrey~E Hinton,
\newblock ``Imagenet classification with deep convolutional neural networks,''
\newblock in {\em Advances in neural information processing systems}, 2012, pp.
  1097--1105.

\bibitem{vedaldi2015matconvnet}
Andrea Vedaldi and Karel Lenc,
\newblock ``Matconvnet: Convolutional neural networks for matlab,''
\newblock in {\em Proceedings of the 23rd ACM international conference on
  Multimedia}. ACM, 2015, pp. 689--692.

\bibitem{wang2004image}
Zhou Wang, Alan~C Bovik, Hamid~R Sheikh, and Eero~P Simoncelli,
\newblock ``Image quality assessment: from error visibility to structural
  similarity,''
\newblock {\em IEEE transactions on image processing}, vol. 13, no. 4, pp.
  600--612, 2004.

\bibitem{sheikh2005information}
Hamid~R Sheikh, Alan~C Bovik, and Gustavo De~Veciana,
\newblock ``An information fidelity criterion for image quality assessment
  using natural scene statistics,''
\newblock {\em IEEE Transactions on image processing}, vol. 14, no. 12, pp.
  2117--2128, 2005.

\end{thebibliography}
\end{document}